\documentclass[10pt]{article}

\usepackage{graphicx}
\usepackage{amsfonts,amssymb,amsmath,amsthm}
\usepackage[sort]{natbib}
\usepackage{algorithm,algorithmic}
\usepackage[hyphens]{url}
\usepackage[pagebackref,letterpaper=true,colorlinks=true,pdfpagemode=none,urlcolor=blue,linkcolor=blue,citecolor=blue,pdfstartview=FitH]{hyperref}
\usepackage{pdfsync}
\usepackage[all]{xy}
\usepackage[margin=1in]{geometry}
\newtheorem*{proposition*}{Proposition}
\newtheorem{theorem}{Theorem}%[section]
\newtheorem{definition}[theorem]{Definition}
\newtheorem{lemma}[theorem]{Lemma}
\newtheorem{corollary}[theorem]{Corollary}
\newtheorem{proposition}[theorem]{Proposition}
\def\indep{\perp\!\!\!\perp}
\newcommand{\given}{\mbox{ }\middle\vert\mbox{ }}

\newcommand{\F}{\mathcal{F}}
\renewcommand{\P}{\mathbb{P}}
\newcommand{\R}{\mathbb{R}}
\newcommand{\X}{\mathcal{X}}

\newcommand{\Y}{\mathcal{Y}}
\newcommand{\TrueMeasure}{\nu}
\newcommand{\RademacherVariable}{\xi}
\newcommand{\RademacherComplexity}{\mathfrak{R}}
\newcommand{\LawOf}[1]{\mathcal{L}\left( #1 \right)}
\DeclareMathOperator*{\argmin}{argmin} % thanks, wikipedia!
\newcommand{\Expect}[1]{\mathbb{E}\left[ #1 \right]}
\newcommand{\Expectwrt}[2]{\mathbb{E}_{ #1 }\left[ #2 \right]}
\newcommand{\ghost}{\widetilde{Z}}
\newcommand{\one}{\mathbf{1}}
\newcommand{\norm}[1]{\left\lVert #1 \right\rVert}
\def\polylog{\operatorname{polylog}}

\usepackage{xspace}
\makeatletter
\newcommand*{\iid}{%
    \@ifnextchar{.}%
        {i.i.d.}%
        {i.i.d.\@\xspace}%
}
\makeatother

\newcommand{\E}{\mathbb{E}}

\usepackage[showonlyrefs]{mathtools}
\newcommand{\email}[1]{\href{mailto:#1}{#1}}

\usepackage{color}
\newcommand{\attn}[1]{\textcolor{red}{#1}}

\title{Rademacher Complexity of Stationary Sequences}

\author{%
 Daniel J.\ McDonald \\
 Department of Statistics\\
 Indiana University\\
 \email{dajmcdon@indiana.edu}
 \and
 Cosma Rohilla Shalizi \\
 Department of Statistics \\
 Carnegie Mellon University\\
 \email{cshalizi@cmu.edu} }

\date{Version: \today}

\begin{document}

\maketitle

\begin{abstract}
  We show how to control the generalization error of time series models
  wherein past values of the outcome are used to predict future values. The
  results are based on a generalization of standard \iid concentration
  inequalities to dependent data without the mixing assumptions common
  in the time series setting. Our proof and the result are simpler
  than previous analyses with dependent data or stochastic adversaries
  which use sequential Rademacher complexities
  rather than the expected Rademacher complexity for \iid
  processes. We also derive empirical Rademacher results
  without mixing assumptions resulting in fully calculable upper bounds.
\end{abstract}

\section{Introduction}
\label{sec:introduction}

Statistical learning theory aims to bound the out-of-sample
performance of prediction rules induced from finite data sets.  The classical
situation is where one wishes to predict one variable $Y \in \Y$ from another
$X \in \X$, and has a training set of $n$ pairs $(X_1, Y_1), \ldots (X_n,
Y_n)$, assumed to be drawn \iid from a distribution $\TrueMeasure$ that will
also generate future instances.  Provided with a loss function $\ell: \Y \times
\Y \rightarrow \R^+$ and a class of prediction functions $\mathcal{G}$, where each $g
\in \mathcal{G}$ is a map from $\X$ to $\Y$, the usual goal is to bound the supremum of
the empirical process of the losses, $\sup_{g \in
  \mathcal{G}}{\Expectwrt{\TrueMeasure}{\ell(Y,g(X))} -
  n^{-1}\sum_{i=1}^{n}{\ell(Y_i,g(X_i))}}$.  Such bounds involve some notion of
the flexibility or complexity of the model space $\mathcal{G}$, and a particularly
important one is the Rademacher complexity,
\[
\RademacherComplexity_n(\mathcal{G}) =
2\Expectwrt{\mathbf{\RademacherVariable},\TrueMeasure}{\sup_{g\in\mathcal{G}}{\frac{1}{n}\sum_{i=1}^{n}{\RademacherVariable_i
        \ell(Y_i,g(X_i))}}}
\]
where $\RademacherVariable_i$ are a sequence of \iid variables taking the
values $+1$ and $-1$ with equal probability (see \S \ref{sec:radem-compl} for a
fuller statement).  While the Rademacher complexity was first used to bound
generalization error for \iid processes~\citep[e.g.][]{Bartlett-Mendelson-on-Rademacher-complexity}, it has been extended to situations
where the $(X_i, Y_i)$ pairs are dependent but $(X_i, Y_i)$ becomes independent
of $(X_j,Y_j)$ as $|i-j| \rightarrow \infty$
\citep{Mohri-Rostamizadeh-rademacher-for-non-iid}, and even to adversarial
settings, where the data source actively tries to fool the learner
\cite{Rakhlin-Sridharan-Tewari-online-learning}.

We build on the latter work to extend Rademacher complexity to the rather
different problem of time-series forecasting.  In that setting, we observe a
single sequence of random variables $Y_1, Y_2, \ldots Y_n$ (for short,
$Y_1^n$), taking values in $\Y$, and wish to learn a function which
extrapolates the sequence into the future, to forecast (say) the
next\footnote{Going beyond ``one-step-ahead'' forecasting, to longer horizons
  or whole blocks, involves mostly notational changes, which we will not note
  explicitly.} value $Y_{n+1}$.  Given a predictor $g: \Y^n \mapsto \Y$, a
natural notion of generalization error for time series, the {\bf forecasting
  risk}, is
%\begin{equation*}
%  \label{eqn:rough-generalization-error}
$R(g) \equiv \Expect{\ell(Y_{n+1},g(Y_1^n))\given Y_1^n}.$
%\end{equation*}

While a precise statement needs some care (\S
\ref{sec:careful-risk-definitions}), we will show that forecasting risk, like
the generalization error of classification and regression problems, can be
bounded via the Rademacher complexity, despite the rather different nature of
the problem. In particular, we are able to use the standard Rademacher
complexity.  Our result is comparable to that in \citet[\S
9]{Rakhlin-Sridharan-Tewari-sequential-complexities}, which gives a bound
for time series prediction in Banach spaces.  Their result however is a
consequence of results for the more difficult problem of prediction under
stochastic adversaries. As such, our bound and the proof are simpler
and tighter, though
they apply to an easier (but still highly relevant) prediction task.

\S\ref{sec:preliminaries} gives background material essential for stating
our results, on time series, model complexity, and forecasting risk.
\S\ref{sec:results} derives risk bounds for time series, giving a novel proof
that the standard Rademacher complexity characterizes the flexibility of
$\mathcal{G}$, even under stationarity, with concentration inequalities for
non-mixing dependent variables.
% \S \ref{sec:examples} supplies some
% straightforward examples showing how dependence affects the quality of
% bounds.
\S\ref{sec:discussion} carefully compares our results to others in the literature,
sketches applications and algorithms, and concludes.

\section{Time Series, Complexity, and Concentration of Measure}
\label{sec:preliminaries}

We introduce some of the concepts needed for our results: stationarity and
ergodicity are required to control generalization error (unless we aim to
predict only a single new observation); Rademacher complexity measures the
flexibility of the model space $\mathcal{G}$; forecasting risk measures the
quality of a time-series prediction rule.

\paragraph{Notation} $\mathbf{Y}=\{Y_t\}_{t=-\infty}^\infty$ is a sequence of
random variables, i.e., each $Y_t$ is a measurable mapping from some
probability space $(\Omega, \mathcal{F}, \mathbb{P})$ into a measurable space
$\Y$.  We write $Y_i^j$ for the block $\{Y_t\}_{t=i, i+1, \ldots j}$ from the
random sequence; either limit may be infinity.  The $\sigma$-field generated by
the block $Y_i^j$ is $\F_i^j$.  $\LawOf{W}$ denotes the probability law of the
random object $W$, and $\LawOf{W|V}$ the conditional law of $W$ given $V$.
Finally, if $W$ has distribution $\nu$ and $f$ is a measurable
function, we define
$\Expectwrt{\nu}{f(W)}=\Expectwrt{W}{f(W)} = \int{d\nu f(W)}$. We will try to
use whichever notation is clearest in context, sticking to $\Expect{f(W)}$ when
that is unambiguous.

\subsection{Stationarity and Ergodicity}
\label{sec:dependent-data}

We assume $\mathbf{Y}$ is (strictly or strongly) stationary.

\begin{definition}[Stationarity]\label{def:stationary}
  A random sequence $\mathbf{Y}$ is {\bf stationary} when all its finite-dimensional
  distributions are time invariant: for all $t$ and all $i \geq 0$,
  $\LawOf{Y_t^{t+i}} = \LawOf{Y_{0}^{i}}$.
\end{definition}
Stationarity does {\em not} require the random variables $Y_t$ to be
independent across time, but does imply they all have the same distribution.

The infinite-dimensional distribution of $\mathbf{Y}$, $\LawOf{\mathbf{Y}}$, is
a probability measure on $\Y^{\infty}$.  In this space, the time-evolution of
the process is just the shift map $\tau$, which ``moves the
sequence a step to the right'': $(\tau \mathbf{Y})_t = \mathbf{Y}_{t+1}$.
\begin{definition}[Ergodicity]
  A set $A \subset \Y^{\infty}$ is {\bf shift-invariant}, $A \in \mathcal{I}$,
  when $\tau^{-1} A = A$.  A probability measure $\mu$ on $\Y^{\infty}$ is {\bf
    ergodic} when shift-invariant sets have either probability 0 or probability
  1, i.e., $A \in \mathcal{I}$ only if $\mu(A) = 0$ or $\mu(A) = 1$.
\end{definition}
Ergodicity is important for two reasons.  The first is that it
implies a law of large numbers for time series.
\begin{proposition}[Individual Ergodic Theorem;
  \citealt{Gray-ergodic-properties-2nd}]
  \label{prop:birkhoff}
  If $\mu$ is stationary and ergodic, and $f \in L_1$, then the time-average of
  $f$ converges to its expectation $\mu$-almost-surely.  That is, the set of $y
  \in \Y^{\infty}$ such that
  $
  \frac{1}{n}\sum_{t=0}^{n-1}{f(\tau^{t} y)} \rightarrow \Expectwrt{\mu}{f(Y)}
  $
  has $\mu$-measure 1.
\end{proposition}
The second reason is that every stationary process $\mathbf{Y}$ decomposes into
a mixture of stationary and ergodic processes, and each realization of
$\mathbf{Y}$ comes from just one of these ergodic components.
\begin{proposition}[Ergodic Decomposition; \citealt{Gray-ergodic-properties-2nd,Dynkin-suff-stats-and-extreme-points}]
  If $\rho$ is a stationary but not ergodic distribution on $\Y^{\infty}$, then
  $\rho = \int{\mu d\pi(\mu)}$, where $\pi$ is a measure on the space of
  stationary and ergodic processes.  Moreover, for any $f \in L_1$,
  $
  \frac{1}{n}\sum_{t=0}^{n-1}{f(\tau^{t} y)} \rightarrow
  \Expectwrt{\rho}{f(Y)|\mathcal{I}}
  $
  for $\rho$-almost-all trajectories $y$.
\end{proposition}
In words, to generate a trajectory from a stationary, non-ergodic process,
first pick a stationary ergodic process (according to the distribution $\pi$),
and then generate $\mathbf{Y}$ from that process.

To sum up, then, if we assume that the data source is stationary, and that we
only get to see a single trajectory from it, there is no loss of generality in
also assuming that the source is ergodic, and so the strong law of large
numbers, in the form of Prop.\ \ref{prop:birkhoff}, applies.  Non-ergodicity
would only be relevant if we were to consider multiple independent trajectories
from the same stationary process, which might sample different ergodic
components \cite{Wiener-prediction}.

\subsection{Empirical Processes}
\label{sec:empirical-processes}

The standard device in learning theory for bounding the generalization error of
a prediction function is to control the {\bf empirical process} over a function
space, i.e., the deviations of empirical means from their expectation values.
We thus define some convenient, if abstract, notation here.

Let $Z_1, \ldots Z_n$ be a sequence of $\mathcal{Z}$-valued random variables
(generally dependent), and $\mathcal{H}$ a class of real-valued functions on
$\mathcal{Z}$.  We define the {\bf empirical mean} or {\bf sample mean} as
$
\hat{h}_n \equiv \frac{1}{n}\sum_{t=1}^{n}{h(Z_t)}
$
and the expectation value as
\[
\Expect{h} \equiv \Expectwrt{Z_1^n}{\hat{h}_n} =
\frac{1}{n}\sum_{t=1}^{n}{\Expectwrt{Z_t}{h(Z_t)}}
\]
If the $Z_t$ are \iid, then $\Expect{h} = \Expectwrt{Z_1}{h(Z_1)}$.  The
empirical process\footnote{Some authors would include an over-all
  scaling factor of $\sqrt{n}$.} at $h$ is
$
\gamma_n(h) = \Expect{h} - \hat{h}_n.$
We care particularly about the supremum of the empirical process:
$
\Gamma_n(\mathcal{H}) \equiv \sup_{h\in\mathcal{H}}{\gamma_n(h)}
$
%%% This is a nice bit, but wastes space here
% (Using ``gamma'' reminds us that when $\Gamma_n(\mathcal{H}) \rightarrow 0$,
% $\mathcal{H}$ is a Glivenko-Cantelli class.)

\subsection{Rademacher Complexity}
\label{sec:radem-compl}

The Rademacher complexity of a function class is, in essence, how well it can
(seem to) match pure noise.  The formal definition is (after
\citealt{Bartlett-Mendelson-on-Rademacher-complexity}):

\begin{definition}[I.i.d.\ Rademacher Complexity]
  \label{def:iid-rademacher}
  Let $Z_1^n$ be a $\Y$-valued \iid sequence, and $\mathcal{H}$ a real-valued
  class of functions on $\Y$.  The {\bf empirical Rademacher complexity} of
  $\mathcal{H}$ on $Z_1^n$ is\footnote{Some definitions have an absolute value inside the
    supremum
    after~\citep{Bartlett-Mendelson-on-Rademacher-complexity}, but
    others avoid it (even the same authors in later work, e.g.~\citealt{BartlettBousquet2005}). As the eventual
    proof demonstrates, it isn't required, so we drop it.}
  \[
  \widehat{\RademacherComplexity}_n(\mathcal{H}) \equiv
  \Expectwrt{\mathbf{\RademacherVariable}}{\sup_{h\in\mathcal{H}}{
      \frac{2}{n}\sum_{t=1}^{n}{\RademacherVariable_t h(Z_t)}}}.
  \]
  The {\bf Rademacher complexity} of $\mathcal{H}$ is the expectation of the
  empirical Rademacher complexity over $Z$:
  $
  \RademacherComplexity_n(\mathcal{H}) \equiv
  \Expectwrt{\mathbf{Z}}{\widehat{\RademacherComplexity}_n(\mathcal{H})}.$
  % =
  % \Expectwrt{\mathbf{Z},\mathbf{\RademacherVariable}}{\sup_{h\in\mathcal{H}}{
  %     \frac{2}{n}\sum_{t=1}^{n}{\RademacherVariable_t h(Z_t)}}}
  % \]
\end{definition}

Rademacher complexity matters because it is closely related to the supremum of
the empirical process over $\mathcal{H}$.  Specifically,
$\Expectwrt{\mathbf{Z}}{\Gamma_n(\mathcal{H})} \leq
\RademacherComplexity_n(\mathcal{H})$. Its utility is that
$\Expectwrt{\mathbf{Z}}{\Gamma_n(\mathcal{H})}$ is almost never expressible, but one of
$\widehat{\RademacherComplexity}_n(\mathcal{H})$ of
$\RademacherComplexity_n(\mathcal{H})$ may be, thus allowing control of
the generalization error with meaningful quantities. The main burden of our paper is to show
that, if $\mathbf{Z}$ is stationary and ergodic rather than \iid, we have the
same result, though with a more involved proof. We rehearse the (now
standard) \iid Rademacher generalization error bound
and its proof using our notation in the Supplement because of its
importance for our own development.

This definition of \iid Rademacher complexity will, it turns out, work for
stationary processes almost unchanged.

\begin{definition}[Rademacher Complexity]
  \label{def:rademacher}
  Let $Y_1^n$ be a time series generated from $\P$.  The
  \emph{empirical Rademacher complexity} of the real-valued function
  class
  $\mathcal{H}$ on $Y_1^n$ is
  \[
  \widehat{\RademacherComplexity}_n(\mathcal{H}) \equiv
  \Expectwrt{\mathbf{\RademacherVariable}}{\sup_{h\in\mathcal{H}}{\frac{2}{n}\sum_{t=1}^{n}{\RademacherVariable_t
        h_t(Y_1^t)}}}
    \]
    The \emph{Rademacher complexity} is the expectation of the empirical
    Rademacher complexity:
  $
  \RademacherComplexity_n(\mathcal{H}) \equiv
  \Expectwrt{Y_1^n}{\widehat{\RademacherComplexity}_n(\mathcal{H})}
  $.
\end{definition}
The term inside the supremum, $\frac{1}{n} \sum_{t=1}^{n}{\RademacherVariable_t
  h_t(Y_1^t) }$, is the sample covariance between the noise
$\RademacherVariable$ and the values of a particular function sequence $h$.
The Rademacher complexity takes the largest value of this sample covariance
over all models in the class (mimicking empirical risk minimization), then
averages over realizations of the noise.

Relative to the \iid Rademacher complexity, we have indexed the predictor $h$
with a time dependent subscript. For time series, the goal is to forecast
$Y_{t+1}$ from the history $Y_1^t$.  Since a function $\Y^t \mapsto Y$ is not,
technically, the same as a function $\Y^{t+1} \mapsto Y$, one must, strictly
speaking, use a different prediction function at each time step.  A single
predictive model $h$ thus is implemented as a whole series of functions $h_t:
\Y^t \mapsto \Y$.  At some abuse of notation, we will write $h$ for the name of
this whole sequence of functions.\footnote{If we wanted to be purists, we could
  introduce a parameter space $\Theta$ (not necessarily finite-dimensional),
  and consider the collection of prediction functions $g_t(Y_1^t;\theta)$ for
  all $t$.} We emphasize that the sequence $h_1,h_2,\ldots$ does not represent
infinitely many individually-learnable functions but rather stages of a
single function sequence $h$.

Intuitively, Rademacher complexity shows how well our models could seem to fit
outcomes which were really just noise, giving a baseline against which to
assess over-fitting or failing to generalize.  Since $\mathbf{Y}$ is stationary
and ergodic, and $\mathbf{\RademacherVariable}$ is \iid and independent of
$\mathbf{Y}$, the joint process $(\mathbf{Y},\mathbf{\RademacherVariable})$ is
also stationary and ergodic.  Thus by the ergodic tower property
\citep{Handel-ergodicity},
% asymptotic weak-mixing of $\mathbf{Y}$ is a sufficient condition such
% that,
for a {\em
  fixed} function sequence $h$, the sample covariance tends to zero
almost surely:
\[
\frac{1}{n}\sum_{t=1}^{n}{\RademacherVariable_t h_t(Y_1^t)}
\rightarrow
\Expectwrt{\mathbf{Y},\mathbf{\RademacherVariable}}{\RademacherVariable
  h(Y)} =
\Expectwrt{\mathbf{\RademacherVariable}}{\RademacherVariable}\Expectwrt{\mathbf{Y}}{h(Y)}
= 0
\]
The overall Rademacher complexity should also shrink, though more slowly,
unless the model class is so flexible that it can fit absolutely anything, in
which case we can infer nothing about how well it will predict in the
future from the fact that it performed well in the past.

Showing that this heuristic reasoning is valid, and that the Rademacher
complexity of Definition \ref{def:rademacher} continues to control the
empirical process when forecasting stationary time series, is the main aim of
our paper.  We note that \citet{Kuznetsov-non-stationary,Kuznetsov-non-mixing}
prove generalization error bounds for the forecasting risk under
non-stationarity with and without mixing assumptions, but these results rely
on the intricate sequential complexities introduced by
\citet{Rakhlin-Sridharan-Tewari-online-learning}, which replace the outer
expectation over the observations $Y_1^n$ with a supremum over such
observations.  (\S \ref{sec:relat-betw-our} carefully compares these results
and ours.)

\subsection{Forecast Risk}
\label{sec:careful-risk-definitions}

In classification or regression problems, we obtain data points $Z_t = (X_t,
Y_t)$, and the goal is to predict one part of the data, $Y_t$, from the other,
$X_t$.  The {\bf risk} of a prediction function $g: \X \mapsto Y$, can be
sensibly defined as an expectation over data points: $ R(g) =
\Expectwrt{X,Y}{\ell(Y,g(X))}.  $ This risk is well-defined so long as the
marginal distribution of the data is shift-invariant ($\LawOf{Z_t} =
\LawOf{Z_1}$ for all $i$).  For an \iid data source, it is of course true that
\begin{align*}
\Expectwrt{X_{n+1},Y_{n+1}}{\ell(Y_{n+1},g(X_{n+1})) \given
  X_1^n,Y_1^n}
& = \Expectwrt{X_{n+1},Y_{n+1}}{\ell(Y_{n+1},g(X_{n+1}))} = R(g)
\end{align*}
so that averaging over the marginal distribution of the next data point
indicates the expected loss of continuing to use the predictor $g$ on new data.
This is no longer true for dependent data.  However, for a stationary ergodic
source, one has that \citep{CRS-Leo-predictive-mixtures}
\begin{align*}
\lim_{m\rightarrow\infty}{\frac{1}{m}\sum_{i=1}^{m}{\Expectwrt{X_{i+n},Y_{i+n}}{\ell(Y_{n+i},g(X_{n+i}))
  \given X_1^n, Y_1^n}}} &= \Expectwrt{X,Y}{\ell(Y,g(X))} = R(g)
\end{align*}
so the expectation over new data would still be a good indicator of long-run
performance.

All of this is subtly changed for time series, where the goal is to forecast
$Y_{t+1}$ from the history $Y_1^t$. As discussed above, we abuse notation and
denote a single predictive model $g$ even though it really represents a of
functions $g_t: \Y^t \mapsto \Y$.

\begin{definition}[Forecast risk]
  Given a stationary and ergodic stochastic process $\mathbf{Y}$, and a
  loss function $\ell$, the \textbf{finite-history risk} of the predictive model
  $g$ is
  \[
  R_n(g) \equiv
  \Expectwrt{Y_1^n}{\frac{1}{n}\sum_{t=1}^{n}{\ell(Y_t,g_t(Y_1^{t-1}))}}
  \]
  and the \textbf{forecast risk} is
  $
  R(g) = \lim_{n\rightarrow\infty}{R_n(g)}
  $
  when the limit exists.
\end{definition}
For brevity, we introduce the notation $Z_t \equiv (Y_t, Y_1^{t-1})$, and
$h_t(Z_t) \equiv \ell(Y_t,g_t(Y_1^{t-1}))$, defining $Y_1^0\equiv\emptyset$.  The forecast risk can thus be also
written as $\lim{n^{-1}\sum_{t=1}^{n}{h_t(Z_t)}}$.
% By ergodicity,\attn{{\scriptsize No, need AEP already here, but then maybe ergodicity gives invariance of limit under conditioning}}
% \[
% \lim_{m\rightarrow\infty}{\frac{1}{m}\sum_{s=1}^{m}{\Expect{h_{n+s}(Z_{n+s}) \given Y_1^n}}} = R(g)
% \]
So $R(g)$, again, captures the long-run average cost of using the predictive
model $g$.  By contrast, $R_n(g)$ is the average risk of $g$ if used on an {\em
  independent} realization of $Y_1^n$.

Having an infinite-time limit in the definition of forecast risk is irksome.
It can be evaded if the predictive model has only a finite memory length $d\geq
0$, so nothing more than $d$ time steps old matters for predictions (formally,
$g_t(Y_1^t) \in \sigma(Y^t_{t-d})$ for all $t > d$).  Then, by stationarity,
we may simplify $ R(g) = \Expectwrt{Y_{1}^{d+1}}{\ell(Y_{d+1},g(Y^d_1))} $ In
fact, in the finite-memory-length case, as soon as $n > d$,
\[
R(g) = \frac{1}{n-d}\sum_{t=d+1}^{n}{\Expectwrt{Y_1^n}{\ell(Y_{t+1},g(Y^{t-1}_{t-d}))}} = R_n(g)
\]
and it follows from the ergodic theorem that
$
\frac{1}{n-d}\sum_{t=d+1}^{n}{\ell(Y_{t+1},g(Y^{t-1}_{t-d}))} \rightarrow R(g)
$ almost surely.

Predictive models with infinite-range memories are however actually fairly
common in forecasting practice, including not just hidden Markov models but
also things as basic as moving-average models.  We therefore {\em posit}
that $R(g)$ exists for such models, writing the gap between the forecast
risk and the finite-history risk by $\Delta_n(g) \equiv R(g) - R_n(g)$.  We
also posit\footnote{If the loss function is the negative log-likelihood,
  this posit is the generalized asymptotic equipartition property, or
  Shannon-McMillan-Breiman theorem
  \cite{Algoet-and-Cover-on-AEP,Gray-entropy}.} that the time-averaged loss
converges to the forecast risk: $ \frac{1}{n}\sum_{t=1}^{n}{h_t(Z_t)}
\rightarrow R(g).  $ With finite amounts of data, we thus focus on control
of $R_n(g)$. Whether $\Delta_n(g)\rightarrow 0$
is a property of both the
function class and the dependence structure, and is outside our scope,
though see \citet{Handel-ergodicity} for related discussion.

\section{Risk Bounds}
\label{sec:results}

Generalization error bounds follow from deriving high probability upper bounds
on the quantity
\begin{equation*}
 \Gamma_n(\mathcal{H}) := \sup_{h \in \mathcal{H}}{R_n(h) - \widehat{R}_n(h)},
\end{equation*}
which is the worst case difference between the true risk $R_n(h)$ and the
empirical risk $\widehat{R}_n(h)$ over all functions in the class of losses
$\mathcal{H} = \{ h = \ell(\cdot, g(\cdot)) : g \in \mathcal{G} \}$ defined
over a particular class of prediction functions $\mathcal{G}$.  We
first present our main result, which bounds $\Expectwrt{\mathbf{Z}}{
  \Gamma_n(\mathcal{H})}$ with the Rademacher complexity and discuss
its proof. We then use
our Rademacher bound to derive risk bounds for time-series forecasters
which are fully calculable from data.

\subsection{Stationary Rademacher Bounds}
\label{sec:stat-radem-bounds}

The symmetrization arguments used to prove Rademacher bounds for the \iid case
fail for time series prediction.  However, as we now show, for stationary time
series, bounds of the same form are still valid, albeit with a somewhat more
involved proof.  This is in contrast to the far more intricate constructions
needed to establish bounds using generalized Rademacher complexities for online
learning
\citep{Rakhlin-Sridharan-Tewari-online-learning,Rakhlin-Sridharan-Tewari-constrained-adversaries}
or for non-stationary processes \citep{Kuznetsov-non-stationary}.  (We give
more detailed contrasts in \S \ref{sec:relat-betw-our}.)   Our
first principle result is simply:
%%%%
% Previous draft of opening paragraph, mine for material in sec:relat-betw-our
% To prove Rademacher bounds, the standard symmetrization argument for the \iid
% case does not work, but, for time series prediction (as opposed to the more
% general dependent data case or the online learning case), Rademacher bounds are
% still available. We emphasize that this result is far simpler than those given
% in either the online learning case
% \citep{Rakhlin-Sridharan-Tewari-online-learning,Rakhlin-Sridharan-Tewari-constrained-adversaries}
% or the non-stationary setting~\citep{Kuznetsov-non-stationary}. Since the
% process is stationary (and not adversarial) our result uses an expectation over
% the sequence $Y_1^n$ rather than a supremum, and our proof is simpler, removing
% the need for constructing a tree-valued process. It is thus analogous to the
% \iid Rademacher complexity, though with a more involved proof.
%%%
\begin{theorem}
  \label{thm:radem}
  For a time series prediction problem based on a sequence $Y_1^n$,
  \[
    \Expect{\Gamma_n(\mathcal{H})} \leq \RademacherComplexity_n(\mathcal{H}).
  \]
\end{theorem}

We note here that unless $\sup_{h\in\mathcal{H}} \norm{h}_\infty<\infty$,
$\RademacherComplexity_n(\mathcal{H})=\infty$ by its definition.  Thus, this
result, like all Rademacher results, is only useful with bounded predictors or
losses.  Of course if $\sup_{h\in\mathcal{H}} \norm{h}_\infty=\infty$, the
theorem holds trivially.

The standard proof for \iid classification or regression introduces a ``ghost
sample'', an independent sample of size $n$ from the same distribution that
produced the original data, before using a symmetrization argument.  For
forecasting, however, introducing an independent copy of the original time
series will not produce the necessary symmetry.  Rather, following an idea
introduced by \citet{Rakhlin-Sridharan-Tewari-online-learning,
  Rakhlin-Sridharan-Tewari-constrained-adversaries} for dealing with
adversarial data, we work with a {\bf tangent sequence}, where the surrogate
value introduced at each time point is conditioned on the actual time series up
to that point.  That is, the tangent sequence $\mathbf{\widetilde{Y}}$ is
defined recursively: $\LawOf{\widetilde{Y}_1} = \LawOf{Y_1}$, and
$\LawOf{\widetilde{Y}_t|Y_1^{t-1}} = \LawOf{Y_t|Y_1^{t-1}}$.  Furthermore,
$\widetilde{Y}_t$ is independent of all other $\widetilde{Y}_s$ and of all
$Y_{s}$, conditional on $Y_1^{t-1}$. (In directed graphical models terms,
$Y_1^{t-1}$ are the parents of $\widetilde{Y}_t$, which has no children.   See \autoref{fig:tree}.)  The
time series $\mathbf{Y}$ and the tangent sequence do {\em not} have the same
joint distributions.\footnote{Let $Y_1$ be $0$ or $1$ with equal probability,
  and $Y_{t+1} = Y_t$ with probability $0.9$ and $= 1-Y_{t}$ otherwise.
  ($\mathbf{Y}$ is a stationary and ergodic Markov chain.)  Because
  $\widetilde{Y}_1 \indep \widetilde{Y}_2 | Y_1$, the probability that
  $\widetilde{Y}_2 = \widetilde{Y}_1$ is not $0.9$ but $0.5$.}

\begin{proof}[Proof of Thm.~\ref{thm:radem}]
With both the original time series $\mathbf{Y}$ and the tangent sequence
$\mathbf{\widetilde{Y}}$ in hand, we construct $Z_t$ and $\ghost_t$ variables
as follows:
$Z_t \equiv (Y_t, Y_1^{t-1})$ and $\ghost_t \equiv (\widetilde{Y}_t, Y_1^{t-1})$
(with the convention that $Z_1 = Y_1$, $\ghost_1 = \widetilde{Y}_1$).  Notice
that $\ghost$ combines the original time series and its tangent sequence,
but in such a way that $\LawOf{Z_t} = \LawOf{\ghost_t}$.  Furthermore,
since $\widetilde{Y}_t\indep Y_t^n|Y_1^{t-1}$, it follows that
$\ghost_t\indep Z_t^n|Z_1^{t-1} $. 
As
$
R_n(h) 
  %=\Expectwrt{\mathbf{Y}}{\frac{1}{n}\sum_{t=0}^{n-1}{\ell(Y_{t+1},g_t(Y_1^{t-1}))}}
  = \Expectwrt{\mathbf{Z}}{\frac{1}{n}\sum_{t=1}^{n}{h_t(Z_t)}} 
  %=\frac{1}{n}\sum_{t=1}^{n}{\Expectwrt{Z_t}{h_t(Z_t)}},
$
for some $h \in \mathcal{H} = \ell \circ \mathcal{G}$,  we may equally
well write the risk in terms of the tangent sequence:
$R_n(h)  =  \frac{1}{n}\sum_{t=1}^{n}{\Expectwrt{\ghost_t}{h_t(\ghost_t)}}$.
Therefore,
  \begin{align}
    \Expect{\Gamma_n(\mathcal{H})} &= \Expectwrt{\mathbf{Z}}{\sup_{h \in\cal{H}}
      \left(\E_{\mathbf{Z}}\left[\frac{1}{n}\sum_{i=1}^n
          h(Z_i)\right] - 
        \frac{1}{n}\sum_{i=1}^nh(Z_i)\right) }\nonumber\\
    &=  \Expectwrt{\mathbf{Z}}{ \sup_{h \in\cal{H}}
      \left(\Expectwrt{\mathbf{\ghost}}{\frac{1}{n}\sum_{i=1}^n
      h(\ghost_i)} - 
        \frac{1}{n}\sum_{i=1}^nh(Z_i)\right)}\nonumber\\
    &\leq \Expectwrt{\mathbf{Z},\mathbf{\ghost}} {\sup_{h
        \in\cal{H}} \frac{1}{n}\sum_{i=1}^n h(\ghost_i) - h(Z_i)} &&
    \mbox{(Jensen's inequality)} \nonumber\\
    \label{eq:here3}
    &= \E_{\substack{Z_1\\\ghost_1}}\E_{\substack{Z_2
        |Z_1\\\ghost_2 |\ghost_1}}\cdots \Expectwrt{\substack{Z_n
        |Z_{n-1},\ldots,Z_1\\\ghost_n |\ghost_{n-1},\ldots,\ghost_1}}{\sup_{h
        \in\cal{H}} \frac{1}{n}\sum_{i=1}^n h(\ghost_i) - h(Z_i)} &&
                                                                 \mbox{(Iterated expectation).}
  \end{align}
  Now, due to dependence, Rademacher variables must be introduced
  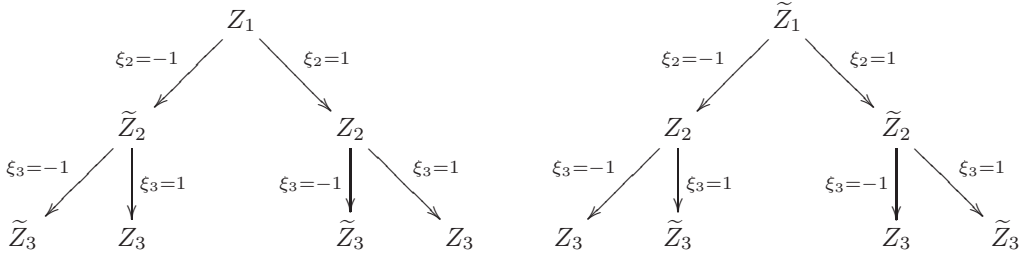
\begin{figure}[t!]
    \centerline{
      \xymatrix{
        &  &   Z_1 \ar[dl]_{\RademacherVariable_2=-1}
        \ar[dr]^{\RademacherVariable_2=1}&&&&& \ghost_1
        \ar[dl]_{\RademacherVariable_2=-1} 
        \ar[dr]^{\RademacherVariable_2=1}\\ 
        &
        \ghost_2\ar[dl]_{\RademacherVariable_3=-1}\ar[d]^{\RademacherVariable_3=1}
        &  
        &Z_2\ar[d]_{\RademacherVariable_3=-1}\ar[dr]^{\RademacherVariable_3=1}&&&
        Z_2\ar[dl]_{\RademacherVariable_3=-1}\ar[d]^{\RademacherVariable_3=1}
        & 
        &\ghost_2\ar[d]_{\RademacherVariable_3=-1}\ar[dr]^{\RademacherVariable_3=1}\\ 
        \ghost_3 &  Z_3 & & \ghost_3  & Z_3& Z_3  & \ghost_3 & & Z_3  & \ghost_3\\
      }
    }   
    \caption{This figure displays the tree structures for
      $\mathbf{Z}(\boldsymbol{\RademacherVariable})$ and
      $\mathbf{\ghost}(\boldsymbol{\RademacherVariable})$ with
      $\RademacherVariable_1=1$ (for example). The path along
      each tree is determined by $\boldsymbol{\RademacherVariable}$ sequence,
      interleaving the ``past'' between paths. The version with
      $\RademacherVariable_1=-1$ would exchange $Z_1$ for $\ghost_1$
      at the root of each tree.} 
    \label{fig:tree}
  \end{figure}
  carefully as in the adversarial case. Rademacher
  variables create two tree structures, one associated to the
  $\mathbf{Z}$ sequence, and one associated to the $\mathbf{\ghost}$
  sequence (see~\citealp{Rakhlin-Sridharan-Tewari-online-learning,Rakhlin-Sridharan-Tewari-constrained-adversaries} for
  a thorough treatment). We write these trees as
  $\mathbf{Z}(\boldsymbol{\RademacherVariable})$ and
  $\mathbf{\ghost}(\boldsymbol{\RademacherVariable})$, where
  $\boldsymbol{\RademacherVariable}$ 
  is a particular sequence of Rademacher variables,
  e.g.~$(1,-1,-1,1,\ldots,1)$, which creates a path along each
  tree. For example, consider
  $\boldsymbol{\RademacherVariable}=\mathbf{1}$. Then,
  $\mathbf{Z}(\boldsymbol{\RademacherVariable}) = (Z_1,\ldots,Z_n)$
  and $\mathbf{\ghost}(\boldsymbol{\RademacherVariable}) =
  (\ghost_1,\ldots,\ghost_n)$, the ``right'' path of both tree
  structures. For $\boldsymbol{\RademacherVariable}=-\mathbf{1}$. Then,
  $\mathbf{Z}(\boldsymbol{\RademacherVariable}) = (\ghost_1,\ldots,\ghost_n)$
  and $\mathbf{\ghost}(\boldsymbol{\RademacherVariable}) =
  (Z_1,\ldots,Z_n)$, the ``left'' path of both tree
  structures. Changing $\RademacherVariable_i$ from $+1$ to $-1$ exchanges $Z_i$
  for $\ghost_i$ in both trees and chooses the left child of $Z_{i-1}$ and $\ghost_{i-1}$
  rather than the right child.  Figure~\ref{fig:tree} displays both
  trees. In order to talk about the probability of $Z_i$ conditional
  on the ``past'' in the tree, we need to know the path taken so
  far. For this, we define a selector function
  $
    \chi(\RademacherVariable) :=
    \chi(\RademacherVariable,\rho,\varrho) =  \rho
    I(\RademacherVariable=1) + \varrho I(\RademacherVariable=-1).
      % \RademacherVariable=1\\ \varrho &
      % \RademacherVariable=-1.\end{cases} 
  $
  Distributions over trees then
  become the objects of interest.
  
  Contrary to the online-learning scenario,
  the 
  dependence between future and past means the adversary is \emph{not} free
  to change predictors and responses separately. Once a branch of
  the tree is chosen, the distribution of future data points is fixed,
  and depends only on the preceding sequence.
  Because of this, the joint distribution of any path along the
  tree is the same as any other path, i.e. for any two paths
  $\boldsymbol{\RademacherVariable},\boldsymbol{\RademacherVariable}'$,
    $\mathcal{L}(\mathbf{Z}(\boldsymbol{\RademacherVariable})) =
    \mathcal{L}(\mathbf{Z}(\boldsymbol{\RademacherVariable'}))$ and
    $\mathcal{L}(\mathbf{\ghost}(\boldsymbol{\RademacherVariable})) =
    \mathcal{L}(\mathbf{\ghost}(\boldsymbol{\RademacherVariable}')).$ 
  Similarly, due to
  the construction of the tangent sequence, we have that
  $\mathcal{L}(\mathbf{Z}(\boldsymbol{\RademacherVariable})) = 
    \mathcal{L}(\mathbf{\ghost}(\boldsymbol{\RademacherVariable}))$. This
    equivalence between paths allows us to introduce Rademacher
    variables swapping $Z_i$ for $\ghost_i$ as well as the ability to
    combine terms below:
  \begin{align*}
    (\ref{eq:here3}) &= \E_{\substack{Z_1\\\ghost_1}}\E_{\RademacherVariable_1}\E_{\substack{Z_2
        |\chi(\RademacherVariable_1,Z_1,\ghost_1)\\\ghost_2
        |\chi(\RademacherVariable_1,\ghost_1,Z_1)}}\E_{\RademacherVariable_2}\cdots
    \E_{\substack{Z_n 
        |\chi(\RademacherVariable_{n-1}),\ldots,\chi(\RademacherVariable_1)\\\ghost_n
        |\chi(\RademacherVariable_{n-1}),\ldots,\chi(\RademacherVariable_1)}}
    \E_{\RademacherVariable_n}\left[\sup_{h
        \in\cal{H}} \frac{1}{n}\sum_{i=1}^n
    \RademacherVariable_i(h(\ghost_i) - h(Z_i))\right]\\ 
    &= \E_{{\mathbf{Z}},\mathbf{\ghost},\RademacherVariable} \left[\sup_{h
        \in\cal{H}} \frac{1}{n}\sum_{i=1}^n \RademacherVariable_i(h(\ghost_i) -
      h(Z_i))\right] \\
    &\leq  \E_{{\mathbf{Z}},\RademacherVariable} \left[\sup_{h
        \in\cal{H}} \frac{1}{n}\sum_{i=1}^n
      \RademacherVariable_ih(Z_i)\right]+\E_{{\mathbf{\ghost}},\RademacherVariable}
    \left[\sup_{h 
        \in\cal{H}} \frac{1}{n}\sum_{i=1}^n \RademacherVariable_ih(\ghost_i)\right]\\
    &= 2 \E_{{\mathbf{Z}},\RademacherVariable} \left[\sup_{h
        \in\cal{H}} \frac{1}{n}\sum_{i=1}^n
      \RademacherVariable_ih(Z_i)\right]= \mathfrak{R}_n(\cal{H}).
  \end{align*}
\end{proof}

Rademacher complexity gets its utility from bounding the prediction risk of
forecasters.  For \iid data, the main tools for proving risk bounds are the
inequalities of \citet{Hoeffding-on-Hoeffding} and
\citet{McDiarmid-on-McDiarmid}.  Extensions of learning theory to dependent
data have relied on strong mixing properties to approximate weakly-dependent
processes by \iid ones, recovering \iid results with a reduced effective sample
size.  We instead use generalizations of Hoeffding and McDiarmid to dependent
sequences based on results of \citet{van-de-Geer-on-Hoeffding-for-dependent},
which do not need mixing at all.  Rather than deriving bounds under the
condition $\sup_{h\in\mathcal{H}} \norm{h}_\infty<\infty$, we use a weaker
hypothesis on the tails of conditional distributions.  We first state this more
general result, giving the bounded case as a corollary.  We discuss the
concentration bound and its derivation in \S\ref{sec:concentration}.

\begin{theorem}
  \label{thm:main}
  Suppose that there exist constants $\tau$ and $c$ such that
  \begin{equation}
    \label{eq:cond-sub-Gaussian}
    \Expect{\psi\left(\frac{|\Gamma_n(\mathcal{H})|}{c}\right) \given
      \F_0^t} \leq \tau \ \ \forall t,
  \end{equation}
  where $\psi(x) = \exp(x^2)-1$.
  Then, for $\epsilon>0$ and $n$ large enough, for all $h\in\mathcal{H}$, with
  probability at least $1-\delta$,
  \[
  R_n(h) \leq \widehat{R}_n(h) + \RademacherComplexity_n(\mathcal{H})+
  4c(\tau+1)\sqrt{\frac{2 \log 1/\delta}{n}}.
  \]
\end{theorem}

The following corollary is immediate by noting that $\sup_{h\in\mathcal{H}}
\norm{h}_\infty \leq M<\infty$ implies that
$\Expect{\exp\left((|\Gamma_n|/M)^2\right)} \leq e<3$. We made no effort to
optimize the constant before the confidence penalty.

\begin{corollary}
\label{cor:bounded-rademacher}
  If $\sup_{h\in\mathcal{H}} \norm{h}_\infty \leq M$, then for all $h\in\mathcal{H}$, with
  probability at least $1-\delta$,
  \[
  R_n(h) \leq \widehat{R}_n(h) + \RademacherComplexity_n(\mathcal{H}) +
  12M\sqrt{\frac{2 \log 1/\delta}{n}}.
  \]
\end{corollary}

\subsection{Empirical Rademacher Bounds}
\label{sec:empir-radem-bounds}

Unfortunately $\RademacherComplexity_n(\mathcal{H})$ may itself be hard or
impossible to calculate for some classes $\mathcal{H}$.  However, under our
assumptions, we show that
$\RademacherComplexity_n(\mathcal{H})$ is closely approximated by the empirical
Rademacher complexity.
That is, the same data can estimate both $\widehat{R}_n$ and
$\RademacherComplexity_n(\mathcal{H})$.

% \begin{theorem}
% \label{thm:radem-convergence}
%   Suppose $\Gamma_n(h)$ satisfies the sub-Gaussian condition in
%   Thm.~\ref{thm:vdg-sub-gaussian} for constants $c,\tau>0$. Then for all $\epsilon>0$,
%   \[
%   \P(\widehat{\RademacherComplexity}_n(\mathcal{H}) -
%   \RademacherComplexity_n(\mathcal{H}) > \epsilon) \leq
%   \exp\left\{-\frac{n\epsilon^2}{128c^2(\tau+1)^2}\right\}.
%   \]
% \end{theorem}

% The following corollary is immediate by applying Thms. \ref{thm:main}
% and \ref{thm:radem-convergence} together with the union bound and
% redefining $\delta\rightarrow \delta/2$.

\begin{theorem}
  [Empirical Rademacher Complexity Bound]
  \label{thm:emp-rad-risk-bound}
  Assume eq.~\eqref{eq:cond-sub-Gaussian} holds. Then, for all $h\in\mathcal{H}$, with
  probability at least $1-\delta$,
  \[
  R_n(h) \leq \widehat{R}_n(h) + \widehat{\RademacherComplexity}_n(\mathcal{H}) +
  12c(\tau+1)\sqrt{\frac{2 \log 2/\delta}{n}}.
  \]
\end{theorem}
% \begin{proof}
%   Apply the union bound to Thms. \ref{thm:main} and \ref{thm:radem-convergence} with
%   $\delta\rightarrow \delta/2$.
% \end{proof}
To apply Thm.~\ref{thm:emp-rad-risk-bound}, we can estimate
$\widehat{\RademacherComplexity}_n(\mathcal{H})$ by drawing $m$
independent Rademacher samples of size $n$, and use
\begin{align}
  \label{eq:estimate-rademacher}
\frac{1}{mn}\sum_{i=1}^m\sup_{h\in\mathcal{H}} \sum_{t=1}^n
\RademacherVariable_{ti}h_t \approx \widehat{\RademacherComplexity}_n(\mathcal{H}).
\end{align}
The approximation is $O(1/m)$-accurate. Thus, given one sample of
data, the entire risk bound is fully calculable. If
$\RademacherComplexity_n(\mathcal{H})$ is known (the case for many
common classes $\mathcal{H}$, see \autoref{sec:examples-rademacher})
we may apply Thm.~\ref{thm:main}. For any other class of predictors, we
can estimate the complexity with \eqref{eq:estimate-rademacher} and
apply Thm.~\ref{thm:emp-rad-risk-bound}. Finally, we present a
corollary for the case that $\sup_{h\in\mathcal{H}} \norm{h}_\infty<\infty$.

\begin{corollary}
\label{cor:emp-rad-risk-bound}
  If $\sup_{h\in\mathcal{H}} \norm{h}_\infty \leq M<\infty$, then for all $h\in\mathcal{H}$, with
  probability at least $1-\delta$,
  \[
  R_n(h) \leq \widehat{R}_n(h) + \widehat{\RademacherComplexity}_n(\mathcal{H}) +
  36M\sqrt{\frac{2 \log 2/\delta}{n}}.
  \]
\end{corollary}

Both of these results can be seen as penalizing
the empirical risk with a term 
that accounts for the complexity of $\mathcal{H}$ along with a second
penalty for the amount of confidence we require.

\subsection{Necessary Concentration Inequalities}
\label{sec:conc-ineq}
\label{sec:concentration}  % Different names at different times...

For \iid data, the main tools for developing risk bounds are the inequalities of
\citet{Hoeffding-on-Hoeffding} and \citet{McDiarmid-on-McDiarmid}.
As discussed above, extensions of learning theory to dependent data have relied on
strong mixing properties to approximate weakly-dependent processes by
\iid ones, and so recover the \iid results with a reduced effective
sample size.  We will instead use a generalization applying to
dependent sequences based on results due to
\citet{van-de-Geer-on-Hoeffding-for-dependent},
which do not require mixing at all.

We need some conditions on the tails of the random variables. Suppose
that $X_t$ is a martingale, e.g.\ a real-valued
$\F_0^t$-measurable random variable satisfying $\Expect{X_t \given
\F_0^{t-1}}=0$ with the convention $\F_0=\varnothing$. For a constant $c$, define
\[
B_n^2 = \sum_{t=1}^n c^2\left( 1+
  \Expect{\psi\left(\frac{|X_t|}{c}\right) \given \F_0^{t-1}}\right),
\]
where $\psi(x) = \exp(x^2)-1$.

Essentially, controlling $B_n^2$ by bounding the expectation of
$\psi(|X_t|/c)$ controls the tails of $X_t$. 
The function $\psi$ can be
any non-decreasing, convex function satisfying
$\psi(0)=0$, but the use of $\psi(x)=\exp(x^2)-1$ is most common. In general,
$\inf_{c>0} \Expect{\psi(|X_t|/c)}\leq 1$ is referred to as the Orlicz
norm of $X_t$ denoted as $\norm{X_t}_\psi$.
In the simplest case,  if $c<\infty$ and $\Expect{X_t}=0$
it holds that $\P(|X_t|>x) \leq 2 \exp(-x^2/c^2)$. This is the
definition of {\bf sub-Gaussian tails}: $X_t$ has tails which decrease
at least as quickly as those of a standard Gaussian random
variable. In particular, bounded random variables satisfy this
condition. As our data come from a time-dependent process
$\mathbf{Y}$, we require the conditional version of this idea.

\begin{lemma}[\citealt{van-de-Geer-on-Hoeffding-for-dependent}; Theorem
  2.2]
\label{lem:vdg-martingale}
  Suppose $X_t$ is a martingale. Then, for all $\epsilon>0,\ b>0$, for $n$ large enough,
  \begin{align*}
  \P\left( \sum_{t=1}^n X_t \geq \epsilon \textrm{ and } B_n^2 \leq b^2\right)
    \leq \exp\{-\epsilon^2/ 8b^2\}.
  \end{align*}
\end{lemma}

This result generalizes Hoeffding's inequality to the case of
conditionally sub-Gaussian random variables from a dependent
sequence. As long as the tails of the next observation are well
controlled \emph{conditional on the past}, we can still control the
size of deviations from the mean with high probability.

We now present the following extension, analogous to McDiarmid's
inequality, but for dependent sequences with sub-Gaussian tails
(rather than bounded differences).

\begin{theorem}
\label{thm:vdg-sub-gaussian}
  Let $X_t$ be $\F_0^t$-measurable with
  \begin{equation}
    \Expect{\psi\left(\frac{|X_t|}{c}\right) \given \F_{0}^{t-1}}
    \leq \tau,
    \label{eq:sub-gaussian}
  \end{equation}
  for some $\tau>0$ and all $t>0$. Then for all
  $\epsilon>0$ and $n$ large enough,
  \[
  \P\left(X_n - \Expect{X_n} > \epsilon\right) \leq
  \exp\left\{-\frac{\epsilon^2}{32nc^2(\tau+1)^2}\right\}.
  \]
\end{theorem}

Thm.~\ref{thm:vdg-sub-gaussian} can be generalized to allow both $c$ and
$\tau$ to depend on $t$ with appropriate modifications as in
Lem.~\ref{lem:vdg-martingale}. Typically, we would expect better control over the tails
as we condition on more data, resulting in a decreasing sequence of
$\tau$, though we will not pursue this generality
further here. Because we were unable to find a comparable result in
the literature, and this one may be useful in it's own right, we have
chosen to include it here. The proof is given in the Supplement.
% Instead, simply for comparison to the \iid case, we
% provide the following simple corollary.

% \begin{corollary}
% \label{cor:vdg-martingale}
%   Let $g(Y_1^n)$ be a real valued function on $\Y^n$ such that
%   \begin{equation}
%     \label{eq:5a}
%     \left|\Expect{g(Y_1^n) \given \F_1^t} - \Expect{g(Y_1^n) \given
%     \F_1^{t-1}}\right| \leq k_t
%   \end{equation}
%   where $k_t$ is $\F_1^{t-1}$-measurable. Then,
%   \begin{align*}
%     \P\left(g(Y_1^n)-\Expect{g(Y_1^n)} >\epsilon \right) &<
%     \exp\left\{-\frac{2\epsilon^2}{\sum_t k_t^2}\right\}.
%   \end{align*}
% \end{corollary}
% In particular, this gives a couple of immediate consequences. If $g$
% is bounded, then
% \begin{equation*}
%     k_t \leq \sup_{Y_t^n}{\sup_{\tilde{Y}_t^n}{\left|
%     g(Y_1^n) -   g(Y_1^{t-1},\widetilde{Y}_t^n) \right|}} =: b_t.
% \end{equation*}
% This contrasts with the bounded differences inequality in the \iid case, wherein
% one only needs to be concerned with one point that is different.  For \iid data,
% starting from Eq.\ \ref{eq:5a} gives
% \begin{equation*}
%     k_t \leq \sup_{Y_t,\tilde{Y}_t}{\left|g(Y_1^n) -
% g(Y_1^{i-1},\tilde{Y}_t,Y^n_{i+1})\right|} =: d_t,
% \end{equation*}
% if $g$ satisfies bounded differences with constants $d_t$.  In other words,
% Corollary \ref{cor:vdg-martingale} combines weak dependence with nice functional
% behavior.

\section{Discussion}
\label{sec:discussion}

In this section, we give a careful explanation, situating our results
in the context of existing bounds. We then provide a few simple
(standard) examples of cases in which our bounds are calculable, as
well as a generalized algorithm for classes which don't admit
calculable expected Rademacher complexities. Finally, we conclude.

\subsection{Relationship with Existing Work}
\label{sec:relat-betw-our}

\begin{table}[tb!]
  \caption{Comparison of existing risk bounds. We use the notation
    $\polylog(n)$ to mean $\log^k (n)$ for some $k>0$.}
  \vskip 0.15in
  \resizebox{\textwidth}{!}{
\begin{small}
\begin{sc}
  \begin{tabular}{ccccc}
    \hline
    Assumptions & Reference & Complexity & Calculable &Best-case
                                                        convergence rate\\
    \hline
    I.i.d. & \citep[many,
             e.g.][]{Bartlett-Mendelson-on-Rademacher-complexity}&
                                                                 Rademacher & Yes & $O(\sqrt{1/n})$\\
    Stationary \& mixing &
                           \citep{Mohri-Rostamizadeh-rademacher-for-non-iid}
                            & Blocked Rademacher & If $\beta$-mixing
                                                   coefs are known &
                                                                     $O(\sqrt{\polylog(n)/n})$\\
    Stationary, non-mixing & This paper & Rademacher & Yes &
                                                             $O(\sqrt{1/n})$\\
    Non-stationary \& mixing &
                               \citep{Kuznetsov-non-mixing,KuznetsovMohri2017}
                            &  Blocked or Sequential Rademacher  & If $\beta$-mixing
                                                                   coefs are known &
                                                                                     $O(\sqrt{\polylog(n)/n})$\\
    Non-stationary, non-mixing & \citep{Kuznetsov-non-stationary} &
                                                                    Expected
                                                                    covering
                                                                    number
                                         & Depends on $\mathcal{H}$
                                                      &
                                                        $O(\sqrt{\polylog(n)/n})$\\
    Adversarial &
                  \citep{Rakhlin-Sridharan-Tewari-constrained-adversaries,Rakhlin-Sridharan-Tewari-online-learning,Rakhlin-Sridharan-Tewari-sequential-complexities}
                            & Sequential Rademacher & Depends on
                                                      $\mathcal{H}$ &
                                                                      $O(\sqrt{\polylog(n)/n})$\\
    \hline
  \end{tabular}
\end{sc}
\end{small}
 }
\label{tab:bound-comparison}
\end{table}

As discussed in the introduction, existing work has developed risk bounds
for dependent data under a number of assumptions which are more or
less general then ours. In order to
give context for our results, we compare the assumptions and benefits
of each of these here. This comparison is summarized in
Table~\ref{tab:bound-comparison}.

The first risk bounds for time series are, like our result, based on standard
Rademacher
complexities. \citet{Mohri-Rostamizadeh-rademacher-for-non-iid} assume
that $\mathbf{Y}$ is a stationary $\beta$-mixing process. Like
our results (Cor.~\ref{cor:bounded-rademacher} and
Cor.~\ref{cor:emp-rad-risk-bound}), they are able to prove bounds based
on both the expected
and empirical Rademacher complexities. Their results however, do not
apply to the full time-series forecasting setting we present
here---predictions in their setting may depend only on a fixed lag $d$
of previous observations. Furthermore,
both the Rademacher
complexity and the confidence penalty depend on blocks of data
rather than individual data points. The number of blocks, $\mu$, then
replaces $n$ in both terms, where $\mu$ depends on the unknown mixing
coefficients. Thus, convergence rates are slightly slower---because the size of
the blocks should increase with $n$, $\mu$ must be sublinear in $n$---and
cannot be directly calculated without knowledge of the mixing
coefficients. \citet{estimating-beta-mixing,McDonaldShalizi2015}
give an estimator for the mixing coefficients with nearly parametric
rates, though bounds which replace known coefficients with estimates have not
been derived. Our results subsume the stationary and
mixing results because our convergence rate is faster without assuming any
type of asymptotic decay of dependence.

Alternatively,
\citet{Rakhlin-Sridharan-Tewari-constrained-adversaries,Rakhlin-Sridharan-Tewari-online-learning,Rakhlin-Sridharan-Tewari-sequential-complexities}
develop truly ingenious techniques for an adversarial data generating process, a much
more general condition wherein not only is the process potentially
non-stationary and non-mixing,
\label{tab:bound-comparison}but subsequent data points may be chosen based on previous predictions
to make the learner perform as poorly as possible. These results rely
instead on the {\bf sequential Rademacher complexity} defined in our
notation as
\[
\RademacherComplexity^{seq}_n(\mathcal{H}) = \sup_{\mathbf{Z}}
\Expectwrt{\mathbf{\RademacherVariable}} {\sup_{h\in\mathcal{H}}{
      \frac{2}{n}\sum_{t=1}^{n}{\RademacherVariable_t h(Z_t(\RademacherVariable))}}},
\]
where the outer supremum is taken over all $\mathcal{Y}$-valued trees
of depth $n$. Because their results are more general, one could simply
apply them to our setting. However,
$\RademacherComplexity^{seq}_n(\mathcal{H})$ is more difficult to
calculate than $\RademacherComplexity_n(\mathcal{H})$, is looser, and
does not admit an empirical version (analogous to our
Cor.~\ref{cor:emp-rad-risk-bound}) because it replaces the outer
expectation over $\mathbf{Z}$ with a supremum.

Finally, work on non-stationary, mixing
processes~\citep{Kuznetsov-non-mixing,KuznetsovMohri2017} and
non-stationary, non-mixing processes~\citep{Kuznetsov-non-stationary}
has also appeared. In the mixing case, the complexity is either the
blocked version as
in~\citep{Mohri-Rostamizadeh-rademacher-for-non-iid} adjusted to
handle non-stationarity, or the sequential complexity above with an
additional discrepancy penalty which ``measures'' non-stationarity in
view of $\mathcal{H}$. The discrepancy measure can be calculated from
data as can the blocked Rademacher complexity, though again, the
mixing coefficients cannot. The non-stationary, non-mixing setting
replaces Rademacher complexities with an {\bf expected sequential
  covering number}. This results in bounds which are looser than ours by
poly-logarithmic factors in $n$. If the covering number can be
computed for the function class $\mathcal{H}$ of interest, than these results are
wholly calculable, but if the class does not have known covering
number, there is no analogue to Cor.~\ref{cor:emp-rad-risk-bound} which
can be estimated from the given data.

Thus, the benefits of our work are that, if we are willing to assume
stationarity, our results are tighter than previous results, easier to
calculate based on known expected Rademacher formulas, and admit empirical
Rademacher complexities which can always be calculated given
sufficient computational resources. None of these benefits require untestable
mixing assumptions or knowledge of the associated coefficients.

\subsection{Examples and Algorithms}
\label{sec:examples-rademacher}

In some cases, the expected (or empirical) Rademacher complexity is easily calculated from
data. In these cases, one can derive simple algorithms for time-series
prediction. Our first two examples, give complete risk bounds for algorithms which
predict future observations based on $d$ previous observations for
clarity. These follow from results of
\citet{Bartlett-Mendelson-on-Rademacher-complexity}.

Consider first the case of a 2-layer Neural
Network which makes predictions based on $d$ previous values and let $\mathcal{Y}=\R^p$. Suppose
that the activation function $\sigma: \R\rightarrow
[-1,1]$ is $1$-Lipschitz with $\sigma(0)=0$. For $v_i\in\R^{pd}$ define
\begin{align*}
\mathcal{G}_N &= \Bigg\{ y \mapsto \sum_i w_i\sigma(v_i \cdot y) :
                %v_i\in \R^d,%\right.\\
            % &\quad\quad\Bigg.
              \norm{w}_1
  \leq 1,\norm{v_i}_1\leq 1 \Bigg\}.
\end{align*}
Suppose further that $\ell$ is $1$-Lipschitz.
Then,
\begin{align*}
\widehat{\RademacherComplexity}_n(\ell \circ \mathcal{G_N}) &\leq
\frac{2c\log^{1/2} (pd)}{n} \max_{1\leq j,j' \leq p}\sqrt{\sum_{i=1}^{n-d} (y_{ij}-y_{ij'})^2}
\end{align*}
for some $c>0$. Thus, $\widehat{\RademacherComplexity}_n(\ell \circ
\mathcal{G_N})= O_{\P}(n^{-1/2})$ as usual. The Lipschitz conditions
and norm constraints can easily be exchanged for other constants
without altering the rate, and the number of layers is easily
altered.

Consider now regularized Kernel methods. Suppose $\ell$ is $M$-Lipschitz and consider the class
$\mathcal{G}_K=\left\{y \mapsto w\cdot \Phi(y): \norm{w}_{\Psi}
  \leq B^2\right\}$, where $\Phi(y): \mathcal{Y} \rightarrow
\Psi$ is the feature map associated with the Hilbert space
$\Psi$, $k$ is the corresponding kernel function, and
$\norm{\cdot}_{\Psi}$ denotes the norm in $\Psi$. Then, we have that
\[
\widehat{\RademacherComplexity}_n(\ell \circ \mathcal{G}_K) \leq
\frac{4MB}{n}\sqrt{\sum_{i=1}^{n-d} k(y_i,y_i)} = O_{\P}(n^{-1/2}).
\]

Finally, using Cor.~\ref{cor:emp-rad-risk-bound}, we can derive a generic
empirical risk minimization-type (ERM)
algorithm for learning without any knowledge of complexity
measurements. Algorithm~\ref{alg:generic} shows how to choose a
predictor from among a collection of bounded function classes
$\mathcal{H}_1,\ldots,\mathcal{H}_k$.

\begin{algorithm}[bt]
  \caption{Generic ERM Algorithm}
  \label{alg:generic}
  \begin{algorithmic}
    \STATE {\bfseries Input:} data $Y_1^n$, models
    $\mathcal{H}_1,\ldots,\mathcal{H}_k$, integer $m$
    \FOR{$i=1$ {\bfseries to} $k$}
    \STATE Estimate a predictor $h_i \in \mathcal{H}_i$ as usual
    \STATE Compute the training error $\widehat{R}_n(h_i)$.
    \STATE Compute $\widehat{\RademacherComplexity}_n(\mathcal{H}_i)$
    using \eqref{eq:estimate-rademacher}
    \ENDFOR
    \STATE Choose $i^*=\argmin_i
  \widehat{R}_n(h_i)+\widehat{\RademacherComplexity}_n(\mathcal{H}_i)$.
  \STATE Return $h_{i^*}$,
  $\widehat{R}_n(h_{i^*})+\widehat{\RademacherComplexity}_n(\mathcal{H}_{i^*})$
  and calculate the complexity penalty to form the bound in
  Cor.~\ref{cor:emp-rad-risk-bound}.
\end{algorithmic}
\end{algorithm}

\subsection{Conclusion}
\label{sec:conclusion}

In this paper, we have demonstrated how to control the generalization of time
series prediction algorithms. These methods use some or all of the observed
past to predict future values of the same series.  In order to handle the
complicated Rademacher complexity bound for the expectation, we have followed
the approach used in the online learning case pioneered by
\citet{Rakhlin-Sridharan-Tewari-online-learning,Rakhlin-Sridharan-Tewari-constrained-adversaries},
but we show that in our particular case, much of the structure needed to deal
with the adversary is unnecessary. This results in clean risk bounds which have
a form similar to the \iid case. As these results take expectations
over $Y_1^n$ rather than a supremum, empirical counterparts which are
estimable can also be derived. Extending our results to local
Rademacher complexities with faster convergence rates is left for
future work.

% The main issue with risk bounds for dependent data is that they rely on
% complete knowledge of the dependence for application. This is certainly true in
% our case in that we need to \emph{know}
% $\Gamma_n(\mathcal{H})$ obeys the conditional Orlicz condition in
% \eqref{eq:cond-sub-Gaussian}.  However, previous results in the dependent data
% setting, such as those presented in
% \citep{Mohri-Rostamizadeh-rademacher-for-non-iid,Karandikar-Vidyasagar-PAC-with-beta-mixing,Mohri-Rostamizdaeh-stability-bounds,Meir-nonparametric-time-series},
% also have this requirement.\footnote{I.i.d.\ results, even more onerously, require
%   us to be able to rule out any dependence at all.} They rely on precise
% knowledge of the mixing behavior of the data which is unavailable. At the same
% time, mixing characterizations are often unintuitive conditions based on
% infinite dimensional joint distributions. Our version depends only on the
% ability to forecastably bound expectations given increasing amounts of data.

\bibliographystyle{mybibsty}
\bibliography{../locusts}

\begin{thebibliography}{999}
\newcommand{\enquote}[1]{``#1''}

\bibitem[Algoet and Cover(1988)Algoet and Cover]{Algoet-and-Cover-on-AEP}
{\sc Algoet, P.~H., and Cover, T.~M.} (1988), \enquote{A sandwich proof of the
  {Shannon}-{McMillan}-{Breiman} theorem,} \emph{Annals of Probability}, {\bf
  16}, 899--909.

\bibitem[Bartlett and Mendelson(2002)Bartlett and
  Mendelson]{Bartlett-Mendelson-on-Rademacher-complexity}
{\sc Bartlett, P.~L., and Mendelson, S.} (2002), \enquote{Rademacher and
  {Gaussian} complexities: Risk bounds and structural results,} \emph{Journal
  of Machine Learning Research}, {\bf 3}, 463--482.

\bibitem[Bartlett et~al.(2005)Bartlett, Bousquet, and
  Mendelson]{BartlettBousquet2005}
{\sc Bartlett, P.~L., Bousquet, O., and Mendelson, S.} (2005), \enquote{Local
  {R}ademacher complexities,} \emph{The Annals of Statistics}, {\bf 33}(4),
  1497--1537.

\bibitem[Dynkin(1978)Dynkin]{Dynkin-suff-stats-and-extreme-points}
{\sc Dynkin, E.~B.} (1978), \enquote{Sufficient statistics and extreme points,}
  \emph{Annals of Probability}, {\bf 6}, 705--730.

\bibitem[Gray(1990)Gray]{Gray-entropy}
{\sc Gray, R.~M.} (1990), \emph{Entropy and Information Theory},
  Springer-Verlag, New York.

\bibitem[Gray(2009)Gray]{Gray-ergodic-properties-2nd}
{\sc Gray, R.~M.} (2009), \emph{Probability, Random Processes, and Ergodic
  Properties}, Springer-Verlag, New York, second edn.

\bibitem[Hoeffding(1963)Hoeffding]{Hoeffding-on-Hoeffding}
{\sc Hoeffding, W.} (1963), \enquote{Probability inequalities for sums of
  bounded random variables,} \emph{Journal of the American Statistical
  Association}, {\bf 58}, 13--30.

\bibitem[Kuznetsov and Mohri(2014)Kuznetsov and Mohri]{Kuznetsov-non-mixing}
{\sc Kuznetsov, V., and Mohri, M.} (2014), \enquote{Generalization bounds for
  time series prediction with non-stationary processes,} in \emph{International
  Conference on Algorithmic Learning Theory}, pp. 260--274.

\bibitem[Kuznetsov and Mohri(2015)Kuznetsov and
  Mohri]{Kuznetsov-non-stationary}
{\sc Kuznetsov, V., and Mohri, M.} (2015), \enquote{Learning theory and
  algorithms for forecasting non-stationary time series,} in \emph{Advances in
  Neural Information Processing Systems 28}, eds. C.~Cortes, N.~D. Lawrence,
  D.~D. Lee, M.~Sugiyama, and R.~Garnett, pp. 541--549.

\bibitem[Kuznetsov and Mohri(2017)Kuznetsov and Mohri]{KuznetsovMohri2017}
{\sc Kuznetsov, V., and Mohri, M.} (2017), \enquote{Generalization bounds for
  non-stationary mixing processes,} \emph{Machine Learning}, {\bf 106}(1),
  93--117.

\bibitem[McDiarmid(1989)McDiarmid]{McDiarmid-on-McDiarmid}
{\sc McDiarmid, C.} (1989), \enquote{On the method of bounded differences,} in
  \emph{Surveys in Combinatorics}, ed. J.~Siemons, pp. 148--188, Cambridge,
  England, Cambridge University Press.

\bibitem[McDonald et~al.(2011)McDonald, Shalizi, and
  Schervish]{estimating-beta-mixing}
{\sc McDonald, D.~J., Shalizi, C.~R., and Schervish, M.} (2011),
  \enquote{Estimating beta-mixing coefficients,} in \emph{Proceedings of the
  $14^{\mathrm{th}}$ International Conference on Artificial Intelligence and
  Statistics [AISTATS 2011]}, eds. G.~Gordon, D.~Dunson, and M.~Dud{\'\i}k,
  vol.~15 of \emph{Journal of Machine Learning Research: Workshops and
  Conference Proceedings}, pp. 516--524.

\bibitem[McDonald et~al.(2015)McDonald, Shalizi, and
  Schervish]{McDonaldShalizi2015}
{\sc McDonald, D.~J., Shalizi, C.~R., and Schervish, M.} (2015),
  \enquote{Estimating beta-mixing coefficients via histograms,}
  \emph{Electronic Journal of Statistics}, {\bf 9}, 2855--2883.

\bibitem[Mohri and Rostamizadeh(2009)Mohri and
  Rostamizadeh]{Mohri-Rostamizadeh-rademacher-for-non-iid}
{\sc Mohri, M., and Rostamizadeh, A.} (2009), \enquote{Rademacher complexity
  bounds for non-{I.I.D.} processes,} in \emph{Advances in Neural Information
  Processing Systems 21 [NIPS 2008]}, eds. D.~Koller, D.~Schuurmans, Y.~Bengio,
  and L.~Bottou, pp. 1097--1104.

\bibitem[Rakhlin et~al.(2010)Rakhlin, Sridharan, and
  Tewari]{Rakhlin-Sridharan-Tewari-online-learning}
{\sc Rakhlin, A., Sridharan, K., and Tewari, A.} (2010), \enquote{Online
  learning: Random averages, combinatorial parameters, and learnability,} in
  \emph{Advances in Neural Information Processing 23 [NIPS 2010]}, eds.
  J.~Lafferty, C.~K.~I. Williams, J.~Shawe-Taylor, R.~S. Zemel, and A.~Culotta,
  pp. 1984--1992, Cambridge, Massachusetts, MIT Press.

\bibitem[Rakhlin et~al.(2011)Rakhlin, Sridharan, and
  Tewari]{Rakhlin-Sridharan-Tewari-constrained-adversaries}
{\sc Rakhlin, A., Sridharan, K., and Tewari, A.} (2011), \enquote{Online
  learning: Stochastic and constrained adversaries,} in \emph{Advances in
  Neural Information Processing Systems 24 [NIPS 2011]}, eds. J.~Shawe-Taylor,
  R.~S. Zemel, P.~Bartlett, F.~C.~N. Pereira, and K.~Q. Weinberger, pp.
  1764--1772.

\bibitem[Rakhlin et~al.(2015)Rakhlin, Sridharan, and
  Tewari]{Rakhlin-Sridharan-Tewari-sequential-complexities}
{\sc Rakhlin, A., Sridharan, K., and Tewari, A.} (2015), \enquote{Sequential
  complexities and uniform martingale laws of large numbers,} \emph{Probability
  Theory and Related Fields}, {\bf 161}(1/2), 111--153.

\bibitem[Shalizi and Kontorovich(2013)Shalizi and
  Kontorovich]{CRS-Leo-predictive-mixtures}
{\sc Shalizi, C., and Kontorovich, A.} (2013), \enquote{Predictive {PAC}
  learning and process decompositions,} in \emph{Advances in Neural Information
  Processing Systems 26}, eds. C.~J.~C. Burges, L.~Bottou, M.~Welling,
  Z.~Ghahramani, and K.~Q. Weinberger, pp. 1619--1627.

\bibitem[van~de Geer(2002)van~de Geer]{van-de-Geer-on-Hoeffding-for-dependent}
{\sc van~de Geer, S.~A.} (2002), \enquote{On {Hoeffding's} inequality for
  dependent random variables,} in \emph{Empirical Process Techniques for
  Dependent Data}, eds. H.~Dehling, T.~Mikosch, and M.~S{\/o}rensen, pp.
  161--169, Birkh{\"a}user, Boston.

\bibitem[van Handel(2014)van Handel]{Handel-ergodicity}
{\sc van Handel, R.} (2014), \enquote{Ergodicity, decisions, and partial
  information,} in \emph{S{\'e}minaire de Probabilit{\'e}s XLVI}, eds.
  C.~Donati-Martin, A.~Lejay, and A.~Rouault, pp. 411--459, Springer.

\bibitem[Wiener(1956)Wiener]{Wiener-prediction}
{\sc Wiener, N.} (1956), \enquote{Nonlinear prediction and dynamics,} in
  \emph{Proceedings of the Third {Berkeley} Symposium on Mathematical
  Statistics and Probability}, ed. J.~Neyman, vol.~3, pp. 247--252, Berkeley,
  University of California Press.

\end{thebibliography}

\clearpage

\appendix

\section{Additional Proofs}
\label{sec:suppl-mater}

  \begin{proposition*}[Standard \iid Rademacher bound]
    If $Z_1,\ldots,Z_n$ is an \iid sample from some probability
    distribution $\P$, then
    $
    \Expectwrt{\mathbf{Z}}{\Gamma_n(\mathcal{H})} \leq \RademacherComplexity_n(\mathcal{H}).
    $
  \end{proposition*}
\begin{proof}
 The usual proof introduces a ``ghost sample'' $\ghost_1^n$, where the
$\ghost_t$ have the same distribution as the $Z_t$, but are independent of
the latter and of each other.  Then expectations may as well be taken over the
ghost sample as the real one:
$
\Expect{h}  =  \Expectwrt{\ghost_1}{h(\ghost_1)}
 =  \frac{1}{n}\sum_{i=1}^{n}{ \Expectwrt{\widetilde{\mathbf{Z}}}{h(\ghost_t)}}.
$
Hence (using the notation from \S \ref{sec:empirical-processes})
\begin{align}
  \gamma_n(h)
  & = \frac{1}{n}\sum_{i=1}^{n}{\Expectwrt{\widetilde{\mathbf{Z}}}{h(\ghost_t)}}
    - \frac{1}{n}\sum_{i=1}^{n}{h(Z_t)}\notag\\
  &=
    \frac{1}{n}\sum_{i=1}^{n}{\Expectwrt{\widetilde{\mathbf{Z}}}{h(\ghost_t)
    - h(Z_t)}},\notag\\
  \Gamma_n(\mathcal{H})
  & \leq
  \Expectwrt{\widetilde{\mathbf{Z}}}{\sup_{h\in\mathcal{H}}{\frac{1}{n}\sum_{i=1}^{n}{h(\ghost_t)
    - h(Z_t)}}}, \label{eq:sup-exp-switch}
\intertext{and}
\Expectwrt{\mathbf{Z}}{\Gamma_n(\mathcal{H})}
& \leq
  \Expectwrt{\mathbf{Z},\widetilde{\mathbf{Z}}}{\sup_{h\in\mathcal{H}}{\frac{1}{n}\sum_{i=1}^{n}{h(\ghost_t)
  - h(Z_t)}}}. \label{eqn:symmetrization}
\end{align}
Eq.~\ref{eq:sup-exp-switch} holds because the supremum of expectations is less than or
equal to the expected supremum, and Eq.\ \ref{eqn:symmetrization} just takes the expectation of
both sides with respect to $Z$.  Since $Z_t$ and $\ghost_t$ have the same
marginal distribution and are independent, $\LawOf{h(\ghost_t) - h(Z_t)} =
\LawOf{h(Z_t) - h(\ghost_t)}$, and the signs of summands in Eq.~\ref{eqn:symmetrization} can be flipped arbitrarily, according to the
Rademacher variables, without effect:
\begin{align*}
\Expectwrt{\mathbf{Z}}{\Gamma_n(\mathcal{H})}
 &\leq
         \Expectwrt{\mathbf{Z},\widetilde{\mathbf{Z}},\mathbf{\RademacherVariable}}
         {\sup_{h\in\mathcal{H}}{\frac{1}{n}\sum_{i=1}^{n}{\RademacherVariable_t\left(h(\ghost_t)
         - h(Z_t)\right)}}}\\
& \leq
  \Expectwrt{\mathbf{Z},\mathbf{\RademacherVariable}}{\sup_{h\in\mathcal{H}}{\frac{1}{n}\sum_{i=1}^{n}{\RademacherVariable_th(Z_t)}}} 
+
  \Expectwrt{\widetilde{\mathbf{Z}},\mathbf{\RademacherVariable}}{\sup_{h\in\mathcal{H}}{\frac{1}{n}\sum_{i=1}^{n}{\RademacherVariable_t
  h(\ghost_t)}}}\\
& =  2
  \Expectwrt{\mathbf{Z},\mathbf{\RademacherVariable}}{\sup_{h\in\mathcal{H}}{\frac{1}{n}\sum_{i=1}^{n}{\RademacherVariable_t
  h(Z_t)}}} = \RademacherComplexity_n(\mathcal{H}).
\end{align*}
\end{proof}

\begin{proof}[Proof of Thm.~\ref{thm:main}]
  This result follows immediately from
  Thm.~\ref{thm:vdg-sub-gaussian} upon setting
    the right hand side equal to $\delta$ and solving for $\epsilon$.
\end{proof}

\begin{proof}[Proof of Thm.~\ref{thm:emp-rad-risk-bound}]
  Write $\mathbf{h}\in \R^n$ for the vector
  $h_1(Z_1),\ldots,h_n(Z_n)$. Note that as the range of $\ell$ is
  $\R^+$, $\mathbf{h}$ lies in the non-negative orthant of $\R^n$
  ($\mathbf{h}\geq 0$). Now,
  \begin{align*}
   n\Gamma_n
   &= \sup_{h\in\mathcal{H}} \left(\sum_{t=1}^n h_t -
     \Expectwrt{\mathbf{Z}}{\sum_{t=1}^n h_t}\right)\\
    &\geq \sup_{h\in\mathcal{H}} \sum_{t=1}^n h_t -
      \sup_{h\in\mathcal{H}} \Expectwrt{\mathbf{Z}}{\sum_{t=1}^n h_t}
   &\textrm{(property of $\sup$)}\\
%    \end{align*}
 %   \begin{align*}
    &\geq \sup_{h\in\mathcal{H}} \sum_{t=1}^n h_t -
     \Expectwrt{\mathbf{Z}}{\sup_{h\in\mathcal{H}} \sum_{t=1}^n h_t}
   &\textrm{(Jensen's ineq.)}\\
   % \end{align*}
   % \begin{align*}
    &= \sup_{\mathbf{h}} \one^\top \mathbf{h} -
     \Expectwrt{\mathbf{Z}}{\sup_{\mathbf{h}} \one^\top \mathbf{h}}\\
    % &\geq \sup_{\boldsymbol{\RademacherVariable}\in \{-1,1\}^n}
    %   \sup_{\mathbf{h}} \boldsymbol{\RademacherVariable}^\top \mathbf{h} -
    %  \sup_{\boldsymbol{\RademacherVariable}\in \{-1,1\}^n}\Expectwrt{\mathbf{Z}}{\sup_{\mathbf{h}} \boldsymbol{\RademacherVariable}^\top
    %   \mathbf{h}}\\
    &\geq \Expectwrt{\boldsymbol{\RademacherVariable}}{
      \sup_{\mathbf{h}} \boldsymbol{\RademacherVariable}^\top \mathbf{h}} -
     \Expectwrt{\mathbf{Z}}{\sup_{\mathbf{h}} \one^\top
      \mathbf{h}}\\
     %  &\geq \Expectwrt{\boldsymbol{\RademacherVariable}}{
    %   \sup_{\mathbf{h}} \boldsymbol{\RademacherVariable}^\top \mathbf{h}} -
    %  \Expectwrt{\boldsymbol{\RademacherVariable}}{\Expectwrt{\mathbf{Z}}{\sup_{\mathbf{h}}
    %     \boldsymbol{\RademacherVariable}^\top
    %   \mathbf{h}}} & (\RademacherVariable\overset{\iid}{\sim}\textrm{Rad.})\\
    %   &= \Expectwrt{\boldsymbol{\RademacherVariable}}{
    %   \sup_{h\in\mathcal{H}} \sum_{t=1}^n\RademacherVariable_t h_t} -
    %  \Expectwrt{\mathbf{Z}}{\Expectwrt{\boldsymbol{\RademacherVariable}}{
    %   \sup_{h\in\mathcal{H}} \sum_{t=1}^n\RademacherVariable_t h_t}}\\
    % &=\frac{n}{2} \left(\widehat{\RademacherComplexity}_n(\mathcal{H})
    %   - \RademacherComplexity_n(\mathcal{H})\right)
   &= \frac{n}{2} \widehat{\RademacherComplexity}_n(\mathcal{H}) - K
  \end{align*}
  where $K$ is a constant. Therefore,
  \[
  \Expectwrt{\mathbf{Z}}{\frac{n}{2}
    \widehat{\RademacherComplexity}_n(\mathcal{H}) - K} =
  \frac{n}{2}\RademacherComplexity_n(\mathcal{H}) - K
  \]
  Since $\psi$ is increasing in it's argument and we assumed that
  $n\Gamma_n$ satisfied eq.~\eqref{eq:cond-sub-Gaussian} for constants
  $c$, and $\tau$, 
  we can apply Thm.~\ref{thm:vdg-sub-gaussian} with $Z_n=
    \widehat{\RademacherComplexity}_n(\mathcal{H}) - K$
    with constants $c\rightarrow 2c/n$ and $\tau$ as before. 
    Thus,
    \[
    \P(\widehat{\RademacherComplexity}_n(\mathcal{H}) -
    \RademacherComplexity_n(\mathcal{H}) > \epsilon) \leq
    \exp\left\{-\frac{n\epsilon^2}{128c^2(\tau+1)^2}\right\}.
    \]
    Setting
    the right hand side equal to $\delta/2$ and combining with
    Thm.~\ref{thm:main} applied with $\delta\rightarrow\delta/2$ via
    the union bound gives the result.
\end{proof}

\begin{proof}[Proof of Thm.~\ref{thm:vdg-sub-gaussian}]
  Write
  $X_n - \Expect{X_n} = \sum_{i=1}^n W_t,$
  where $W_t = \Expect{X_n\given \F_0^t} - \Expect{X_n\given
    \F_0^{t-1}}$, for all $t=1,\ldots,n$. Then $W_t$ is
  $\F_0^t$-measurable, and $\Expect{W_t\given \F_0^{t-1}} = 0$ for
  all $t$. Now, let $K>0$ to be chosen. Then
  \begin{align*}
    \Expect{\psi(|W_t|/K)\given \F_0^{t-1}}
    &=\Expect{\psi\left(\frac{|\Expect{X_n\given \F_0^t} - \Expect{X_n\given
      \F_0^{t-1}}|} {K}\right)\given \F_0^{t-1}}\\
    &=\mathbb{E}\left[\exp\left\{\frac{1}{K^2}\left(\Expect{X_n\given \F_0^t}
      - \Expect{X_n\given
      \F_0^{t-1}}\right)^2\right\}- 1\given \F_0^{t-1}\right]\\
    &\leq \mathbb{E}\left[\exp\left\{\frac{2}{K^2}\left(\Expect{X_n\given \F_0^t}^2
      + \Expect{X_n\given
      \F_0^{t-1}}^2\right)\right\}- 1\given \F_0^{t-1}\right]\\
    &= \mathbb{E}\left[\exp\left\{\left(\frac{|\Expect{X_n \given \F_0^t}|}
      {K/\sqrt{2}} \right)^2\right\} \exp\left\{\left(\frac{|\Expect{X_n \given \F_0^{t-1}}|}
      {K/\sqrt{2}}\right)^2 \right\} -1 \given \F_0^{t-1} \right]\\
    &= \exp\left\{\left(\frac{|\Expect{X_n \given \F_0^{t-1}}|}
      {K/\sqrt{2}} \right)^2\right\} \Expect{\exp\left\{\left(\frac{|\Expect{X_n \given \F_0^t}|}
      {K/\sqrt{2}} \right)^2\right\} \given \F_0^{t-1} }-1\\
  % \end{align*}
  % \begin{align*}
    &\leq \Expect{\exp\left\{\left(\frac{|X_n |}
      {K/\sqrt{2}} \right)^2\right\} \given
      \F_0^{t-1}}\Expect{\Expect{\exp \left\{\left(\frac{|X_n|}
      {K/\sqrt{2}} \right)^2\right\} \given \F_0^t} \given \F_0^{t-1}
      }-1\\
    &= \Expect{\exp\left\{\left(\frac{|X_n |}
      {K/\sqrt{2}} \right)^2\right\} \given \F_0^{t-1}} \Expect{\exp\left\{\left(\frac{|X_n|}
      {K/\sqrt{2}} \right)^2\right\} \given  \F_0^{t-1} }-1\\
    &=\Expect{\exp\left\{\left(\frac{|X_n |}
      {K/\sqrt{2}} \right)^2\right\} \given \F_0^{t-1}}^2 - 1\\
    &\leq (\tau+1)^2-1
  \end{align*}
  for $K=c\sqrt{2}$. Therefore, we have
  \begin{align*}
    B_n^2
    &= \sum_{i=1}^n
            2c^2\left(1+\Expect{\psi(|W_t|/\sqrt{2}c)\given
            \F_0^{t-1}}\right)
    \leq 2nc^2(\tau+1)^2,
  \intertext{and so,}
    \P\left(X_n - \Expect{X_n} > \epsilon\right)
      &=\P\left(\sum_{i=1}^n W_t > \epsilon\right)
    \leq \exp\left\{-\frac{\epsilon^2}{32nc^2(\tau+1)^2}\right\},
  \end{align*}
  by Lem.\ \ref{lem:vdg-martingale}.
\end{proof}

\end{document}